%% file: main.tex
\documentclass[conference]{IEEEtran}
\IEEEoverridecommandlockouts
\usepackage{cite}
\usepackage{amsmath,amssymb,amsfonts}
\usepackage{algorithmic}
\usepackage{graphicx}
\usepackage{textcomp}
\usepackage{xcolor}

\usepackage{bm}
\usepackage{ascmac}
\usepackage{url}
\usepackage{multirow}
\usepackage{orcidlink}
\usepackage[subrefformat=parens]{subcaption}

\makeatletter
\newcommand{\figcaption}[1]{\def\@captype{figure}\caption{#1}}
\newcommand{\tblcaption}[1]{\def\@captype{table}\caption{#1}}
\makeatother

\def\BibTeX{{\rm B\kern-.05em{\sc i\kern-.025em b}\kern-.08em
    T\kern-.1667em\lower.7ex\hbox{E}\kern-.125emX}}
\begin{document}

\title{OpenPack: A Large-scale Dataset \\ for Recognizing Packaging Works \\ in IoT-enabled Logistic Environments}

\author{
\IEEEauthorblockN{Naoya Yoshimura, Jaime Morales, Takuya Maekawa, Takahiro Hara}
    \IEEEauthorblockA{Graduate School of Information Science and Technology, Osaka University, Japan}
    \{yoshimura.naoya, jaime.morales, maekawa, hara\}@ist.osaka-u.ac.jp
}

\maketitle

\begin{abstract}
    \input{body/00_Abstract}
\end{abstract} 


\input{body/10_Introduction}
\input{body/20_Related_work}
\input{body/30_Dataset}
\input{body/40_DatasetAnalysis}

\input{body/71_Evaluation}
\input{body/72_Results}
\input{body/50_Discussion.tex}

\input{body/80_Conclusion}

\vspace{1em}
\noindent
\textbf{Ethical Statement}: 
This study is approved by the ethical committee of the institute of the authors.

\vspace{0.25em}
\noindent
\textbf{Acknowledgement}: 
This work is partially supported by JSPS KAKENHI JP21H03428, JP21H05299, JP21J10059, JST ACT-X JPMJAX200T, and JST Mirai JP21473170. 
We greatly appreciate Chikako Kawabe, Kana Yasuda, and Makiko Otsuka for their efforts in developing the OpenPack dataset.
We would also like to express our appreciation to Dr. Namioka, Toshiba Corporation for the support in data collection.

\bibliographystyle{IEEEtran}
\bibliography{ref}

\end{document}

%% file: body/00_Abstract.tex
Unlike human daily activities, existing publicly available sensor datasets for work activity recognition in industrial domains are limited by difficulties in collecting realistic data as close collaboration with industrial sites is required. This also limits research on and development of methods for industrial applications.
To address these challenges and contribute to research on machine recognition of work activities in industrial domains, in this study, we introduce a new large-scale dataset for packaging work recognition called OpenPack.
OpenPack contains 53.8 hours of multimodal sensor data, including acceleration data, keypoints, depth images, and readings from IoT-enabled devices (e.g., handheld barcode scanners), collected from 16 distinct subjects with different levels of packaging work experience.
We apply state-of-the-art human activity recognition techniques to the dataset and provide future directions of complex work activity recognition studies in the pervasive computing community based on the results.
We believe that OpenPack will contribute to the sensor-based action/activity recognition community by providing challenging tasks.
The OpenPack dataset is available at \url{https://open-pack.github.io}.

%% file: body/10_Introduction.tex
\section{Introduction}

In factories and logistics centers, human workers continue to perform important roles in adapting to the fast-changing demands of customers and suppliers \cite{michel2016,yavas2020logistics}. 
In 2021, Amazon shipped over 5 billion packages in the U.S.\footnote{https://www.seattletimes.com/business/amazon-is-upss-biggest-customer-and-biggest-competitive-threat} with 1.6 million employees\footnote{https://www.macrotrends.net/stocks/charts/AMZN/amazon/number-of-employees}, emphasizing the need for efficient shipping in supply chains.
Digitization is applied in industry to streamline human work and assist in decision-making as part of Industry 4.0. In Industry 4.0, data from sensors and IoT devices (e.g., connected handheld terminals) is used to recognize and streamline human work activities. Therefore, activity recognition for human workers in industry has become a notable research topic in pervasive computing \cite{sozo2019nursing,xia2019unsupervised,niemann2020lara,xia2020robust,morales2022acceleration,yoshimura2022losnet}.

\begin{figure*}[t]
  \centering
  \includegraphics[width=1.00\textwidth]{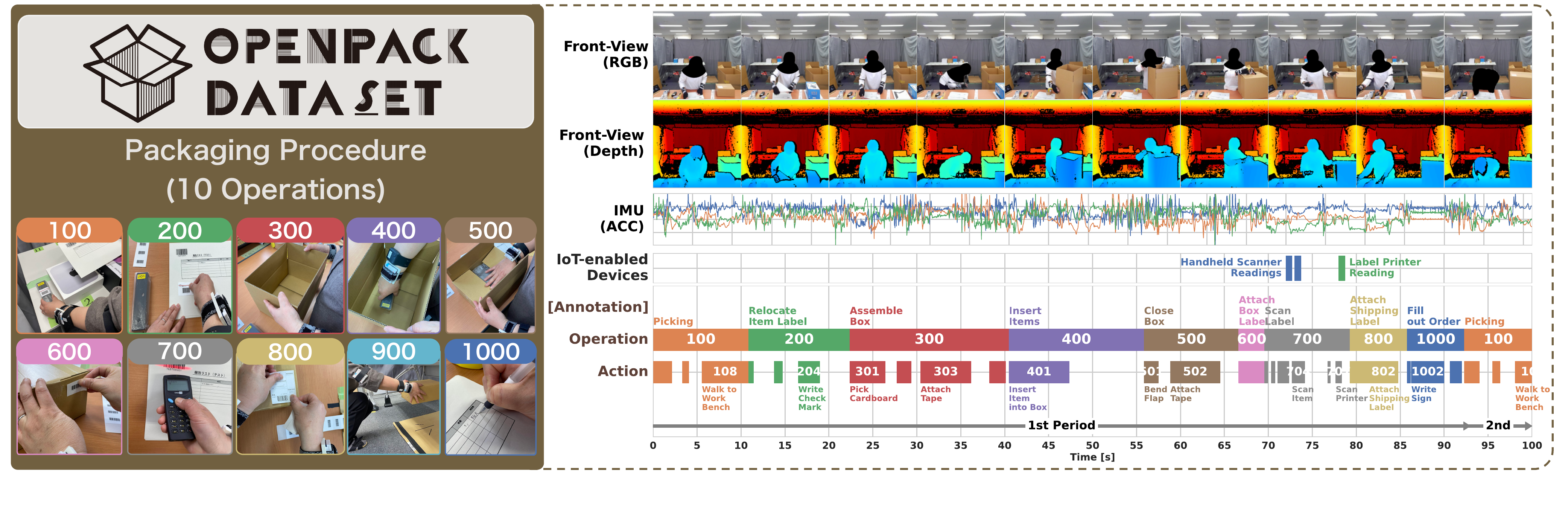}
  \vspace{-10mm}
  \caption{
    Illustration and example sensor data of the OpenPack dataset. 
    A subject iterated a typical series of packaging works, with each iteration of the process (i.e., period) comprising a sequence of complex operations.
  }
  \label{fig:main}
\end{figure*}

However, the following challenges should be addressed to enhance work activity recognition studies.
\newline
\noindent
\textbf{\textbf{$\bullet$}  Lack of datasets for industrial domains}: Fig. \ref{fig:main} shows a typical series of packaging tasks iterated several times, with each iteration of the process (i.e., period) comprising a sequence of operations in which the acceleration data indicate the complexity of operations.
However, the amount of publicly available datasets for industrial domains containing complex activities remains limited, and many of the available activity recognition datasets focus only on simple daily activities (e.g., walking and running) \cite{roggen2010,pamap2Dataset,barshan2014recognizing}. 
\newline
\noindent
\textbf{\textbf{$\bullet$}  Limited modality}: Many public datasets for manual tasks provide only vision-related modalities \cite{tang2019coin,dallel2020inhard,ben2021ikea}. However, because many types of manufacturing equipment and storage systems are installed in industrial environments, occlusion tends to pose challenges in applying vision-only approaches.
\newline
\noindent
\textbf{\textbf{$\bullet$}  Unavailability of readings from IoT-enabled devices}: Although digitization is progressing in actual industrial domains as part of the development of Industry 4.0, to our knowledge, no datasets on activity recognition that include both sensory data on human motions and readings from IoT-enabled devices operated by the human workers are publicly available.
\newline
\noindent
\textbf{\textbf{$\bullet$}  Lack of rich metadata}: Many of the available datasets do not provide a rich set of metadata related to manual works such as a set of the items to be packed, which limits the understanding of recognition results and the design of new, enhanced research tasks.

To address the challenges regarding complex work recognition, in this study, we propose a new multimodal dataset for packaging work recognition in logistics, called OpenPack (Fig. \ref{fig:main}).
OpenPack consists of 20,161 instances of activities (operations) and 53,286 instances of actions with 9 types of modalities captured from 16 distinct subjects with various levels of experience performing packaging tasks.
The main features of OpenPack are summarized as follows.

\noindent \textbf{(1) OpenPack is the largest multimodal work activity dataset in the industrial domain}, including sensory data from body-worn inertial measurement units (IMUs), depth images, and LiDAR point clouds for use in research on multi-/crossmodal, IMU-only, and vision-only work activity recognition according to conditions in an expected target environment.\\
\noindent \textbf{(2) OpenPack provides a rich set of metadata} such as subjects’ levels of experience in packaging work as well as their physical characteristics, enabling the design of various research tasks (e.g., assessment of workers’ performance) in addition to basic work activity recognition.\\
\noindent \textbf{(3) OpenPack is the first large-scale dataset for complex packaging work recognition that contains readings from IoT-enabled devices}. Leveraging high-confidence readings from IoT-enabled devices, which strongly relate to the activities performed (e.g., handheld scanner readings relate to ``scan label'' operations), is expected to be a key enabler to precise recognition in real-world applications.

The key contributions of this research are:\\
\noindent \textbf{$	\bullet$} To the best of our knowledge, OpenPack is the largest dataset on industrial work activity recognition. We expect this data to contribute to the pervasive computing community by providing a challenging task, that is, to perform complex activity recognition via sensor and sparse IoT data.\\
\noindent \textbf{$	\bullet$}  We apply state-of-the-art recognition methods to OpenPack and provide future research directions for complex work activity recognition by analyzing the recognition results.\\

%% file: body/20_Related_work.tex
\section{Related Work}

\setlength{\tabcolsep}{1mm}
\begin{table*}[t]
  \centering
  \caption{
    Overview of public datasets of human activities/works. D: Depth, Acc: Acceleration data, Gyro: Gyroscopic data, Ori: Orientation sensor data, EDA: Electrodermal activity, BVP: Blood volume pulse, Temp: Temperature. 
  }
  \label{tab:datasets}
  {\footnotesize
    \begin{tabular}{l||lp{15mm}lllll|p{25mm}p{13mm}l|c} \hline
                                           &                       & \textbf{Type of}  & \textbf{Recording}      & \textbf{Activity}        & \textbf{\# of}       & \textbf{Annotated} &                   &                                               & \textbf{IoT}        &                    &               \\
      \textbf{Domain}                      & \textbf{Datasets}     & \textbf{Task} & \textbf{Length}         & \textbf{Class$^{3}$}         & \textbf{Work Period} & \textbf{Instances} & \textbf{Subjects} & \textbf{Modality}                             & \textbf{Devices}    & \textbf{View}      & \textbf{Year} \\ \hline \hline
      \multirow{3}{*}{Multi-modal}         & MHAD                  & Daily Actions         & 82m                     & 11                       & N/A                  & 660                & 12                & RGB+D+Keypoints +Acc+Mic                      & No                  & 12                 & 2013          \\ \cline{2-12}
                                           & UTD-MHAD              & Daily Actions         & 9m                      & 27                       & N/A                  & 180                & 8                 & RGB+D+Keypoints +Acc+Gyro                     & No                  & 1                  & 2015          \\  \cline{2-12}
                                           & MMAct                 & Daily Actions         & 17h35m                  & 37                       & N/A                  & 36,764             & 20                & RGB+D+Keypoints +Acc+Gyro+Ori +WiFi+Pressure  & No                  & 4+Ego              & 2019          \\ \hline \hline
      \multirow{2}{*}{Cooking}             & CMU-MMAC              & Cooking               & -                       & 5                        & 186                  & 186                & 39                & RGB+D+Keypoints +Acc+Mic                      & Yes (RFID)          & 5                  & 2010          \\ \cline{2-12}
                                           & 50 Salads             & Cooking               & 4h                      & 52                       & 50                   & 2,967              & 25                & RGB+D+Acc                                     & Yes (Acc)           & 1                  & 2013          \\ \hline \hline
      \multirow{3}{*}{\begin{tabular}{l}Procedural\\Activity\end{tabular}} & COIN                  & Instruction Video     & 476h38m                 & 180                      & N/A                  & 46,354             & N/A               & RGB (Youtube)                                 & No                  & N/A                & 2019          \\ \cline{2-12}
                                           & IKEA-ASM              & Furniture Assembly    & 35h16m                  & 33                       & 371                  & 16,764             & 48                & RGB+D+Keypoints                               & No                  & 3                  & 2021          \\ \cline{2-12}
                                           & Assembly101           & \begin{tabular}{l}Toy\\Assembly\end{tabular}          & 42h+                    & 202                      & 362                  & 1M+                & 53                & RGB                                           & No                  & 12                 & 2022          \\ \hline \hline
      \multirow{5}{*}{Industrial}          & InHARD                & Industrial Actions    & 18h30m                  & 14 + 72                  & 38                   & 4,800               & 16                & RGB+Keypoints (3D)                            & No                  & 3                  & 2020          \\ \cline{2-12}
                                           & ABC Bento             & \begin{tabular}{l}Bento\\Packaging\end{tabular}       & 3h22m                   & 10                       & 199                  & 151                & 4                 & MOCAP                                         & No                  & 1                  & 2021          \\ \cline{2-12}
                                           & \multirow{2}{*}{LARa} & Picking               & \multirow{2}{*}{14h50m} & \multirow{2}{*}{8  + 19} & 324                  & 8,878               & 16                & RGB+Keypoints (3D)                            & \multirow{2}{*}{No} & \multirow{2}{*}{1} & 2020          \\ \cline{3-3} \cline{6-8}
                                           &                       & Packaging             &                         &                          & 125                  & 2,103               & 10                & +Acc                                          &                     &                    & +2022         \\ \cline{2-12}
                                           & OpenPack (v1.0.0)              & Packaging             & 53h50m                  & 10 + 32                  & 2,048                 & 20,161             & 16                & D+Keypoints+LiDAR +Acc+Gyro+Ori +EDA+BVP+Temp & Yes                 & 2                  & 2023          \\ \hline \hline
    \end{tabular}
    \vspace{-2.8mm}
  }
\end{table*}

Although many multimodal, vision-based, and IMU-based datasets for daily activity recognition have been made publicly available \cite{spriggs2009cmummac,ofli2013berkeley,chen2015utd,kong2019mmact}, the number of publicly available datasets for work activity recognition in industrial domains remains limited.
Table \ref{tab:datasets} summarizes the attributes of datasets on human activities and manual labor\footnote{Activity Class: When action classes are defined in a hierarchical manner, the numbers of classes in different levels are shown with separator ``+''. In the case of OpenPack, they correspond to \# of operations and actions.} .
To the best of our knowledge, the LARa dataset \cite{niemann2020lara,niemann2022context} is the only dataset on work activity recognition in logistics.
However, this dataset does not have class labels of types of operations such as packing items and assembling a box, and the subjects has no or limited experience in working at real logistics centers. 
Moreover, the above datasets do not provide readings from IoT-enabled devices or a rich set of metadata.

The InHARD dataset \cite{dallel2020inhard} and the ABC Bento packaging dataset \cite{abc2021bento} are designed to accelerate human-robot collaboration in industrial settings.
The InHARD dataset consists of RGB and 3D keypoint data from 16 subjects collected while they were assembling various parts and components.
The ABC Bento Packaging dataset is a dataset that captures activities related to bento packaging. This dataset focuses on common mistakes made by bento manufacturers, such as ``forgetting to put in ingredients,'' and provides labels only for outlier types.
The ABC Bento Packaging dataset is quite small to be applied to data-driven algorithms, such as deep learning.
These vision-based datasets also lack sensor data modalities.
In contrast, OpenPack is a large-scale dataset containing 20,161 work operation instances with multimodal sensor data.

Datasets of various complex procedural activities are also available.
Specifically, many multimodal/vision-based cooking activity datasets are available such as CMU-MMAC \cite{spriggs2009cmummac}, 50 Salads dataset \cite{stein2013combining}, Breakfast Actions Dataset \cite{kuehne2014language}, EPIC-KITCHENS \cite{damen2018epick}, and the Cooking Activity Dataset \cite{abc2020cook}.
Vision-based datasets focused on procedural activities other than cooking include IKEA-ASM \cite{ben2021ikea} and Assembly101 \cite{sener2022assembly101} for assembling furniture and toys, and COIN \cite{tang2019coin}, a collection of instructional videos collected from YouTube.
As noted above, many multimodal or vision-based datasets on manual tasks in daily life have been made publicly available.
In contrast, the availability of public multimodal datasets for industrial domains remains limited, and OpenPack is the first large-scale dataset for activity recognition in industrial domains.
This may be attributed to the difficulties in collecting datasets for industrial domains compared to everyday tasks.
Collecting data for industrial applications requires close collaboration with industrial engineers working in an actual target environment to coordinate an experimental environment with various equipment for the target task, to define a set of activity labels by obtaining an actual work instruction document used in the target industrial environment, and to employ workers with experience in the target task as research subjects.

%% file: body/30_Dataset.tex
\section{Developing OpenPack Dataset}

OpenPack (\url{https://open-pack.github.io}) is the first multimodal large-scale dataset for activity recognition in industrial domains.
16 distinct participants packed 3,956 items in 2,048 shipping boxes in total, and the total duration of our dataset is 53.8 hours, consisting of 104 data collection sessions.
OpenPack is the largest industrial dataset that includes both vision and wearable sensor data with precise labels by annotators.

In the following, we present an overview of the target activity of the dataset, the process by which it was collected, and how the data were annotated.

\begin{figure}[t]
    \begin{center}
        \includegraphics[width=1.0\linewidth]{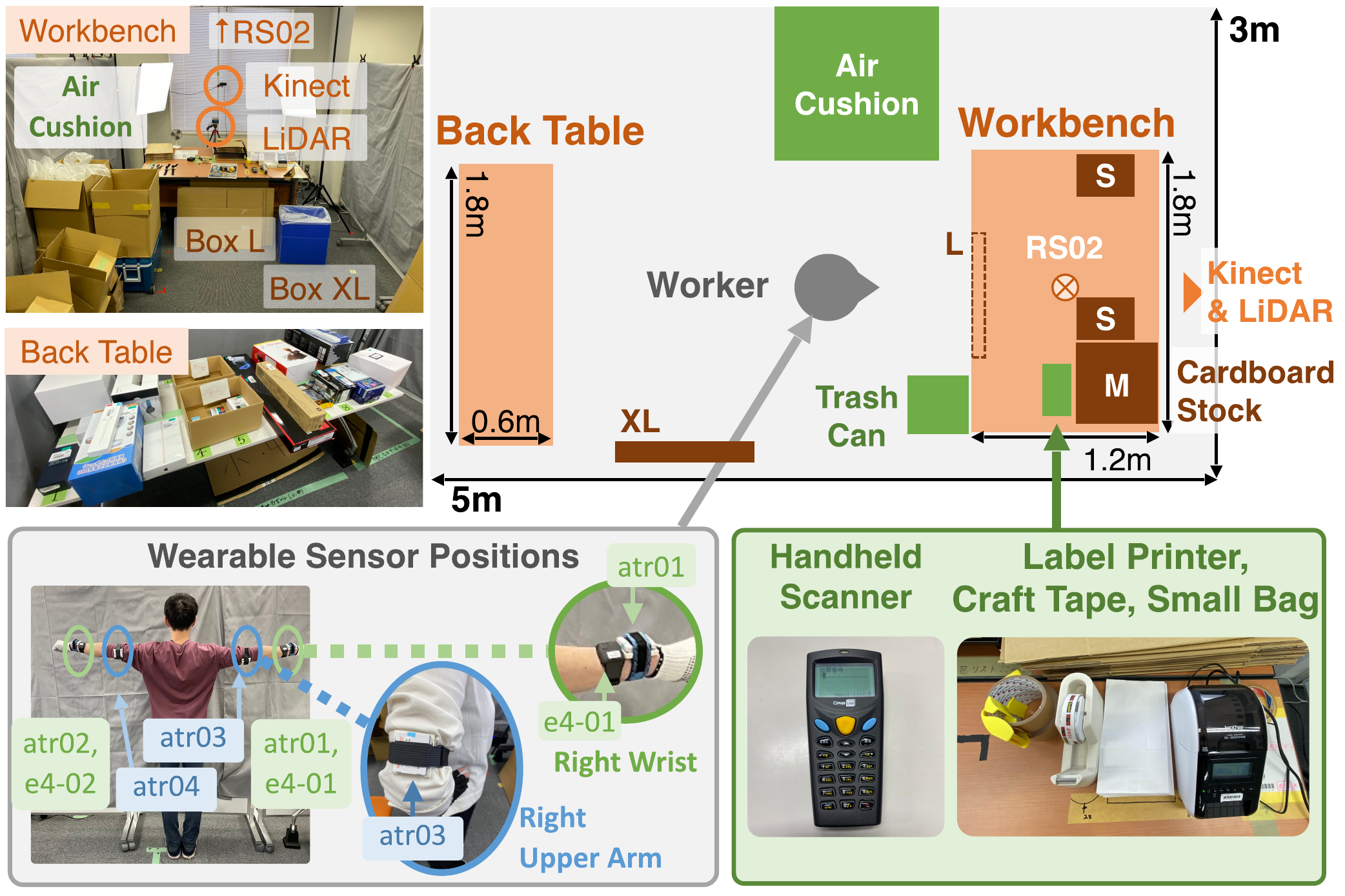}
        \vspace{-5mm}
        \caption{
            Environmental setup and wearable sensor positions
        }
        \label{fig:environment}
    \end{center}
\end{figure}

\subsection{Packaging Work}

As shown in Fig. \ref{fig:main}, a typical series of complex operations is iterated, with each iteration (i.e., period) comprising a sequence of operations such as assembling a shipping box and filling the box with items.
In a given work period, a worker processes a single shipping order consisting of 10 pieces of work.
That is, the worker picks items in the shipping order, double-checks the items, assembles a shipping box, fills the box with the items, and so forth to complete the order.
When performing specific operations, the worker uses IoT-enabled devices such as a handheld barcode scanner, and the operation is recorded and transmitted by the device.

Because the size of items to be packed, the number of items, and the size of shipping items depend on shipping orders, sensor data collected in different work periods and the duration of the same operation in different work periods vary.
The task of recognizing specific operations is challenging owing to these characteristics of packaging work.

\subsection{Data Collection}

\begin{figure*}[t]
    \begin{center}
        \includegraphics[width=0.97\textwidth]{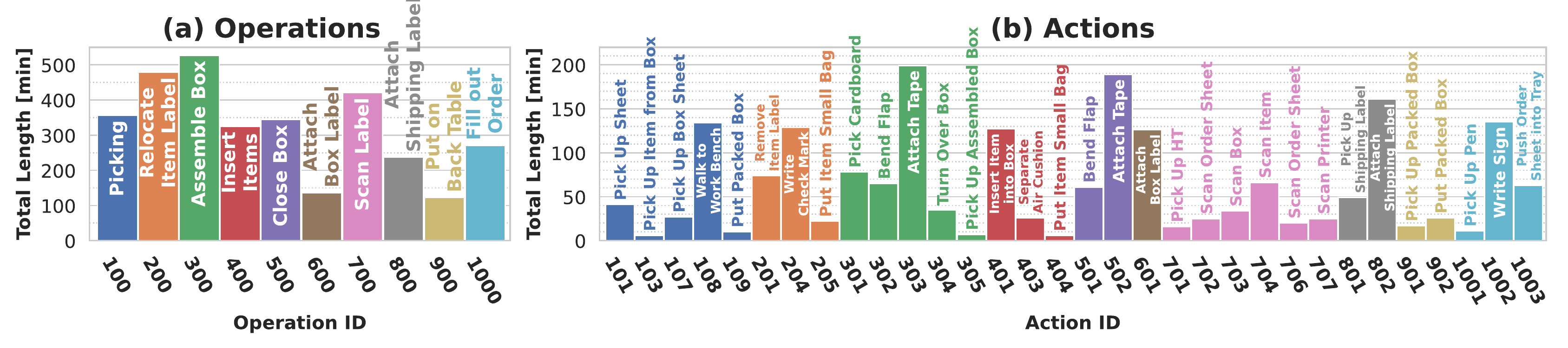}
        \vspace{-3mm}
        \caption{
          Distribution of the total lengths of the annotated activities.
          The horizontal axis shows activity IDs.
        }
        \label{fig:stats_label_dist}
        \vspace{-1.0em}
    \end{center}
\end{figure*}

\subsubsection{Collection Environment}

We collected data in a dedicated environment shown in Fig. \ref{fig:environment}.
With the help of industrial engineers, we constructed a 3m $\times$ 5m environment designed to simulate an actual workspace in a warehouse.
The environment mainly comprised a workbench, a back table on which items were placed after being picked from shelves by another worker, boxes containing air cushions, and a trash can.
A handheld barcode scanner, a printer, and craft tapes used for packaging were located on the right side of the table. Four types of cardboard boxes were available for packing, including small, medium, large, and extra-large sizes.

\subsubsection{Subjects}

We invited 16 subjects (5 males and 11 females) to participate in our data collection process.
The ages of the subjects were ranged from 20s and 50s.
12 subjects had experience in packaging work ranging from 1 month to 4 years.
In addition, 4 subjects did not have work experience.
Data from these four subjects can be valuable for recognizing the operations of workers newly involved in the work environment and analyzing the learning curve. 
One subject was left-handed.
Each subject was assigned a consistent identifier throughout the entire dataset.

\subsubsection{Data Modalities}

OpenPack provides nine data modalities, including acceleration, gyroscope, quaternion, blood volume pulse (BVP), electrodermal activity (EDA) data, body temperature, keypoints, a LiDAR point cloud, and depth images.
Fig. \ref{fig:environment} illustrates the positions of wearable sensors.
Four IMU units were attached to the subject's left and right wrists and upper arms to collect acceleration data on three axes, as well as gyroscope and quaternion data at 30 Hz.
In addition, two Empatica E4 sensors were attached to the subject's left and right wrists to collect BVP and EDA signals at 64 Hz and 4 Hz, respectively, in addition to acceleration data at 32 Hz.
Kinect and LiDAR sensors were installed as front-view cameras and RealSense (RS02) as a top-view camera as shown in Fig. \ref{fig:environment}.
The LiDAR sensor was considered effective in accurately tracking the subject's position when they were away from the workbench.
We included the BVP and EDA sensors because they are expected to be used for analyzing the internal status of subjects in specific situations.
OpenPack also provides operational logs of IoT-enabled devices, i.e., the handheld scanner and label printer, in the environment.
The operational logs of the handheld scanner, for example, contain a time-stamp of a scan and an identifier of a scanned item.
These are highly reliable sources of information for recognizing the scanning operation. %

\subsubsection{Data Collection Procedure}

Before data collection, each subject received instructions related to the outline of the experiment and the operations to be performed based on the instruction document.
Subsequently, we obtained the informed consent of the subjects, who then practiced the packaging work by performing work periods up to five times.
Subsequently, we activated and calibrated the sensors and attached the wearable sensors to the subjects.
The subjects iterated up to five data collection sessions within 6 h (including a 60-min lunch break).
At the beginning of each session, the subject received 20 order sheets and then sequentially processed the sheets.
That is, the subject completed 20 shipping boxes in the session.
A 15-min break was included between two consecutive sessions.

\subsubsection{Scenarios}
\label{subsubsec:scenarios}

The difficulties in packaging work recognition depend on various factors found in logistics centers.
Four scenarios are prepared to simulate the difficulties.
\textbf{Scenario 1}: This is the most simple scenario in which workers follow the work instructions as accurately as possible.
\textbf{Scenario 2}: Some logistics centers do not require workers to strictly follow work instruction documents. %
Here, we encouraged the subjects to alter the procedure of operations for more efficient operation. %
\textbf{Scenario 3}: Depending on the items to be packed and/or surrounding situations, a worker performs irregular actions or activities.
Here, three irregular situations/actions were added: (1) use an already assembled box, (2) put small items into an additional bag, and (3) pick up items for several order sheets at the same time.
\textbf{Scenario 4}: Due to seasonal events or flash sales, task volumes for each worker may temporarily increase and workers have to move more quickly.
Here, we rushed the subjects by introducing an auditory alarm to simulate a busy working time.

\subsubsection{Metadata}

OpenPack provides a rich set of metadata, which is mainly composed of subject- and order-related metadata.
The subject-related metadata contains information regarding each of the participants’ experience in packaging tasks, as well as their dominant hand, gender, and age.

Order-related metadata contains information regarding items and an order sheet processed by a subject in a work period.
OpenPack assumes an online order management system and provides information regarding an identifier of an order and a set of items to pack in the order.
The management system also stores information regarding an identifier, product code, and the size of each item.
These identifiers are used to manage items in an order management system.
In contrast, product codes are unique numbers for each item, which are widely used in retail sales and enable information to be retrieved regarding an item, such as product name, product type, and price.
OpenPack also provides this information.

\subsection{Annotations}

After data collection, the data were labeled by three expert annotators by reference to the RGB images with the help of industrial engineers. 
It took about 1 year to complete the annotation task.
Two types of labels are available: (1) work operation and (2) action.

\subsubsection{Work Operations}

Operation classes used in OpenPack were defined based on an instruction document used in an actual logistics center.
The document specifies a sequence of operations performed by a worker, and each worker in the center performs operations according to the document.
Therefore, the basic activities performed by all workers, i.e., operations, were used to label the dataset.
Our dataset contains ten classes of operations shown in Fig. \ref{fig:stats_label_dist} (a). 
Note that many other logistics centers also utilize patterns of operations very similar to those used in this study.

\subsubsection{Actions}
An instructional document also contains a description of each operation that explains how to perform the operation.
For example, a description of the ``relocate item label'' operation is given as ``\textit{Remove the label from the items and place it on the bottom margin of the packaging list. Check the product name and quantity on the list and label with a ballpoint pen}.''
Based on the description, we also defined 32 action classes included in operations as shown in Fig.~\ref{fig:stats_label_dist} (b).
For example, Fig. \ref{fig:main} shows that the ``assemble box'' operation is composed of four actions, including ``pick up cardboard,'' ``bend flap,'' ``attach tape,'' and ``turn over box.''
The action classes are useful for a manager in a logistics center to assess the status of a job in progress in detail.

Note that action labels were not assigned to every time step, as shown in Fig.~\ref{fig:main}. We did not annotate all the atomic actions included in the operations because there are many meaningless and inconsistent body movements, for example, the transition from one action to the next. Therefore, we created action labels that specified meaningful and consistent atomic actions, as described in the work instruction document.

For the details of the data collection and annotation, see the dataset website (\url{https://open-pack.github.io}).

%% file: body/40_DatasetAnalysis.tex
\section{Dataset Analysis}
\label{sec:dataset_analysis}

In this section, we provide some statistics on the dataset to show the diversity of packaging work in terms of activity length and period length.

\subsubsection{Annotation Summary}

Figure~\ref{fig:stats_label_dist} summarizes the total length for each class.

The total lengths of operation and action labels are 53.8 hours and 37.8 hours, respectively.
The numbers of operation and action labels are 20,161 and 53,286, respectively.
As shown in Fig.~\ref{fig:stats_label_dist}, there is considerable variation in the total lengths of labeled segments in the activities.
This reflects a large difference in the time required to perform different work operations/actions and their occurrence frequency.
To recognize these activities in OpenPack, this class imbalance problem must be considered. This could be challenging, especially for action recognition.

\subsubsection{Variation in Operations and Period Length}
\label{subsubsec:variation_in_Operations}

\begin{figure}[t]
  \centering
  \includegraphics[width=1.0\linewidth]{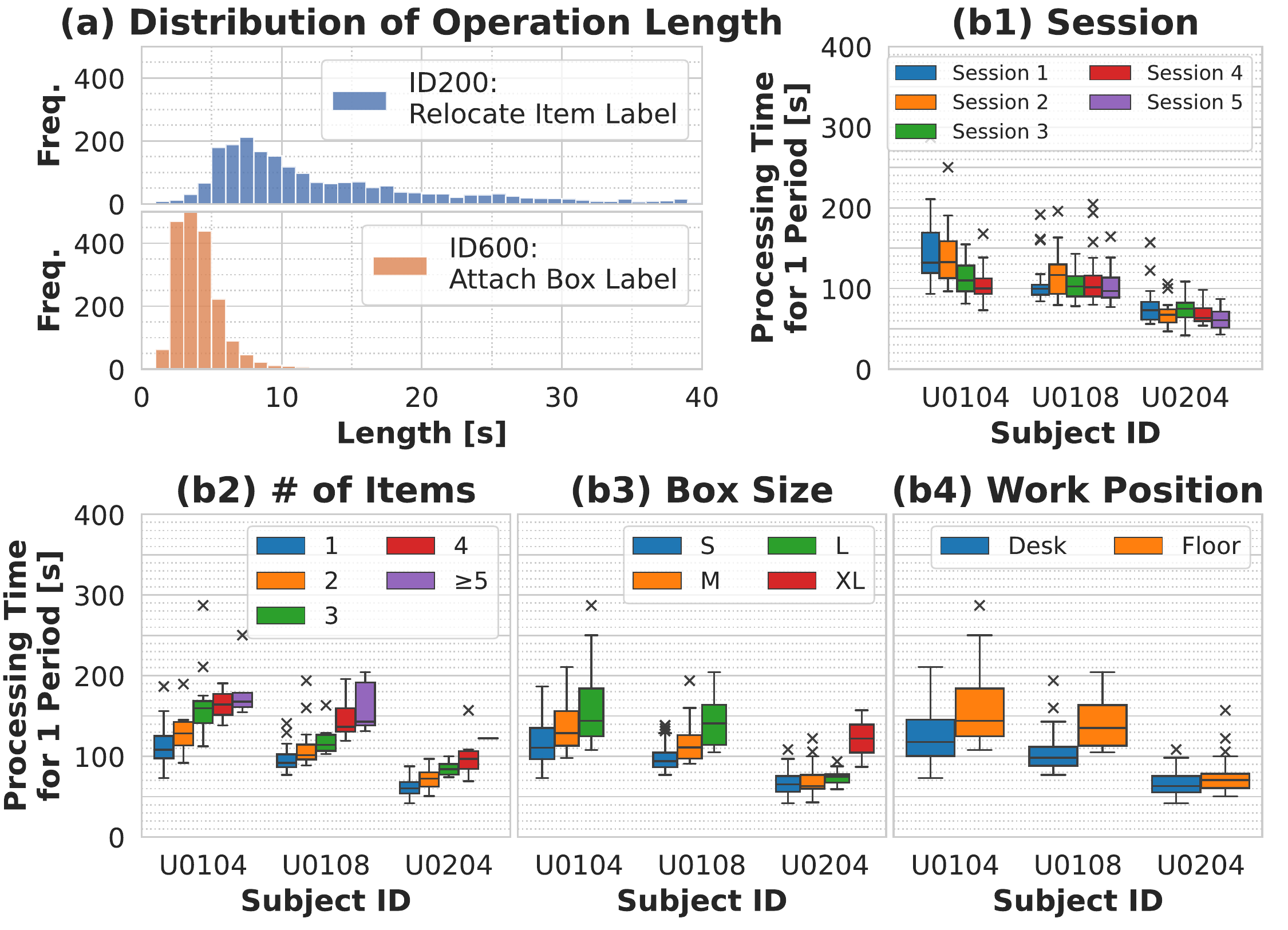}
  \vspace{-5mm}
  \caption{
    (a) Distribution of the length of the two operation classes.
    (b) Distribution of the period lengths with different sessions, item numbers, box sizes, and work positions.
  }
  \label{fig:ds_analysis}
\end{figure}

Even for the same work activities, the data exhibit a wide variation depending on different situations.
Fig.~\ref{fig:ds_analysis}~(a) shows the distribution of the length of two operations; ``relocate item label,'' and ``attach box label.''
Simple operations such as ``attach box label,'' which are not affected by the number of items in an order sheet, only exhibit small variations in length.
In contrast, operations such as ``relocate item label,'' which depend on the numbers and sizes of items, exhibit long-tailed distributions.

Figure~\ref{fig:ds_analysis}~(b1) shows the distribution of period lengths calculated for each user and session.
The working speeds of the subjects vary significantly. For example, the work speed of U0204 is particularly fast; this worker completed a single period in 70.1 seconds on average, while the average for all workers is 96.4 seconds.
The effects of the number of items to be packed, box size, and work location, on the period length are summarized in Fig.~\ref{fig:ds_analysis}~(b2--b4).
As the number of items and box size increases, the period length tends to increase.
In addition, work performed with boxes placed on the floor tended to take longer time than work performed on a workbench, owing to the time required to move items and tools.
Although these three factors are not independent, they have a significant impact on work activities and makes recognition difficult.

%% file: body/71_Evaluation.tex
\section{Evaluation \& Benchmark}
\label{sec:evaluation}
\subsection{Evaluation Methodology}

We benchmarked state-of-the-art activity recognition methods on the OpenPack dataset in the four typical settings.

Acceleration data from the workers’ left wrists, i.e., non-dominant hands, were used as inputs.
Models were trained to recognize 10 work operations at 1 Hz resolution.
Macro average of F1-measure calculated for each scenario was used as evaluation metrics.
We prepared the following 6 models as baselines: CNN \cite{ordonez2016deep}, U-Net \cite{zhang2019human}, DeepConvLSTM (DCL) \cite{ordonez2016deep}, DCL with Self-attention (DCL-SA) \cite{singh2020deep}, ConformerHAR \cite{gulati2020conformer}, and LOS-Net(-R) \cite{yoshimura2022losnet}.
Models are trained with the Adam optimizer with ExponentialLR learning rate scheduler for up to 500 epochs with early stopping.
We repeated the training five times with different random seeds, and the average score is shown.

%% file: body/72_Results.tex
\subsection{Results}
\label{subsec:result}

\subsubsection{Cross-user Activity Recognition with a Sufficient Amount of Training Data (Data-rich Setting)}
\label{subsubsec:benchmark1}

\begin{figure}[t]
  \begin{minipage}[t]{1.0\linewidth}
    \centering
    \includegraphics[width=1.035\linewidth]{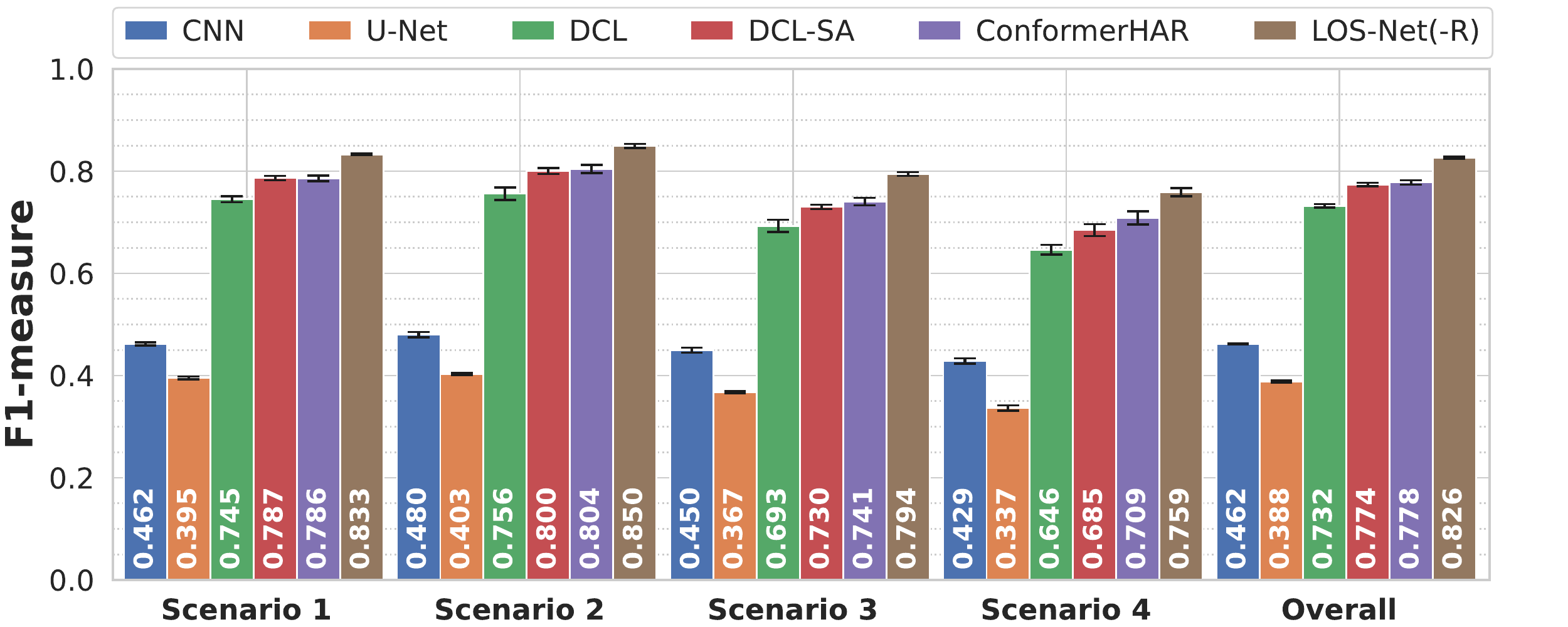}
    \vspace{-5mm}
    \caption{
      Results of cross-user work operation recognition with a sufficient amount of training data
    }
    \label{fig:f1_b1}
  \end{minipage} \\

  \vspace{1.0em}
  \begin{minipage}[t]{1.00\linewidth}
    \centering
    \includegraphics[width=1.035\linewidth]{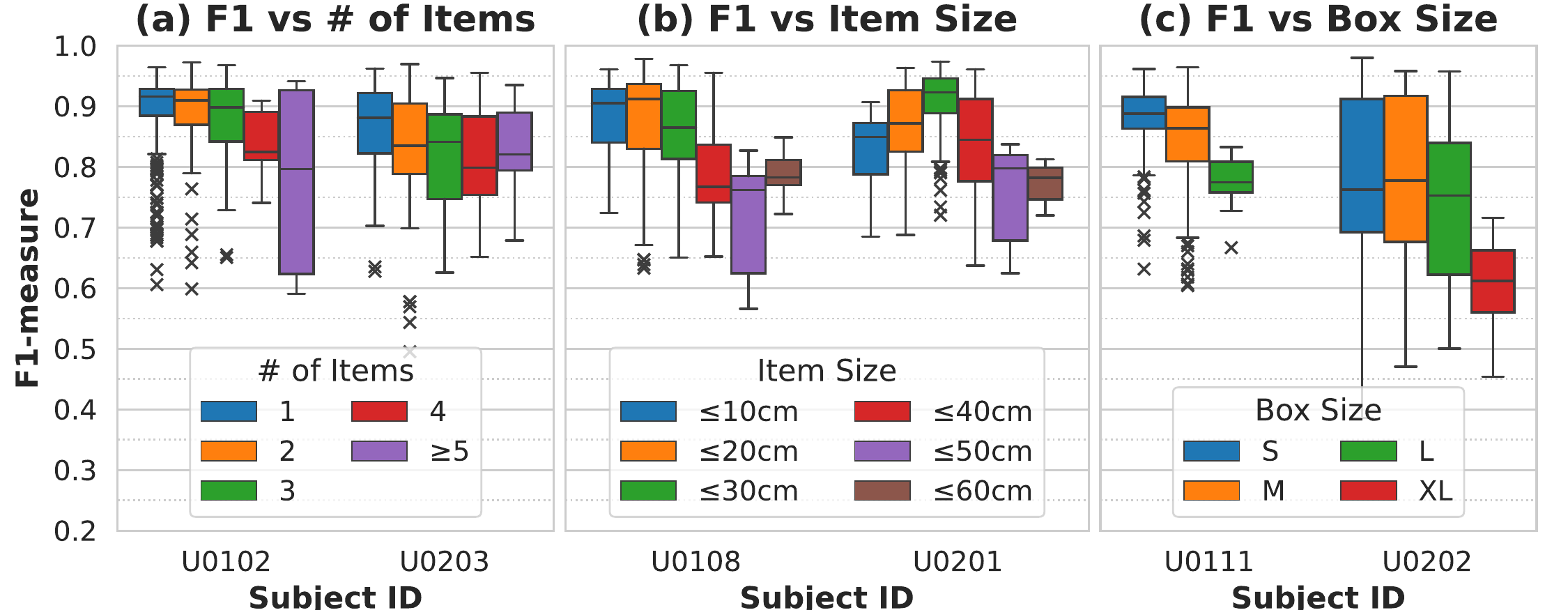}
    \caption{
      Distributions of F1-measure in different conditions (LOS-Net(-R); Cross-user setting)
    }
    \label{fig:f1_b5_period}
  \end{minipage}
\end{figure}

This setting is designed to confirm the upper bounds of recognition performance of complex work activities for deep models that require large amounts of training data.
This data-rich setting assumes that a sufficient amount of training data from other users are provided, i.e., the cross-user setting.
Therefore, we conducted the leave-one-subject-out cross validation.

Figure \ref{fig:f1_b1} shows the results of work operation recognition.
Long-term context must be extracted to estimate the class to which each action belongs based on the previous and next actions because most operations are similar to each other in their movements. 
Thus, LOS-Net(-R), which has a module for long-term context extraction, achieved the highest F1-measures of 0.83 in Scenario 1.
In contrast, CNN and U-Net models perform relatively poorly at extracting long-term contexts, as shown in the experimental results.
DCL-SA slightly outperformed DCL, which demonstrates the effectiveness of the self-attention mechanism on the work activity recognition task.
The scores for Scenario 4 were lower than those of the other scenarios because the subjects were rushed.
In reality, it is difficult to prepare such a large amount of training data like this benchmark setting.
It is necessary to develop a model that can recognize with the same high accuracy even if the amount of training data is limited.

Figure~\ref{fig:f1_b5_period} shows the distributions of F1-measures on the period basis in different conditions: (a) the number of items to be packaged, (b) the size of items, and (c) the size of used boxes. 
The recognition performance decreases with an increase in the item count or size because there are limited samples corresponding to them in the training dataset. 
Specifically, the difference in the item size significantly affects the worker's movements. 
As shown in Fig.~\ref{fig:f1_b5_period} (b), the F1-measure of U0201 for ``$\leq 30$cm'' is higher than that of the other conditions. 
This might be owing to the larger hand movements in ``$\leq 30$cm'', which makes recognition easier.
In reality, the item distribution is also biased, which may degrade the recognition performance.
Developing techniques to mitigate this degradation by utilizing item metadata would be an interesting research topic.

\subsubsection{Work Activity Recognition with a Limited Amount of Training Data from a Known Worker (Data-scarce Setting)}
\label{subsubsec:benchmark2}

When work instruction documents are not very strict, workers perform tasks in different ways, which makes cross-user operation recognition more difficult.
However, preparing a sufficient amount of labeled training data from a target worker is expensive.
Therefore, the objective of this setting was to investigate the performance of recognizing workers' activities with a limited amount of training data collected from a target worker.
In this data-scarce setting, a limited amount of training data (1 session of data) from each target subject was used.
Each model was trained on data from the 3rd session and calculated the F1-measure for each scenario using the remaining data from the same subject.
There are only 20 periods in one session, but the annotation took roughly 5 hours.

\begin{figure}[t]
  \centering
  \includegraphics[width=1.035\linewidth]{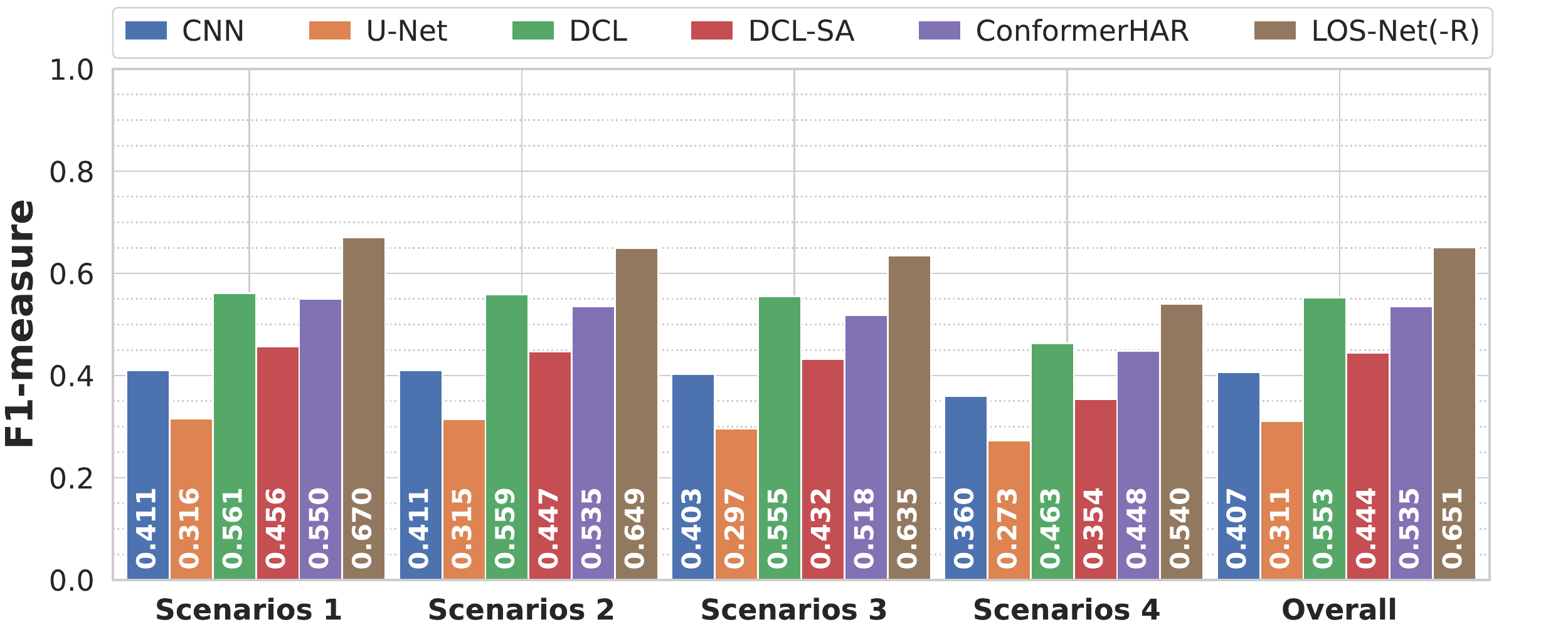}
  \vspace{-5mm}
  \caption{
    Results of work operation recognition with a limited amount of training data from a known worker.
  }
  \label{fig:f1_b2}
\end{figure}

Figure~\ref{fig:f1_b2} shows the results of the average of F1-measure for all users in each scenario.
In Scenario 1, LOS-Net(-R) achieved the F1-measure of 0.67, but they were 0.16 pts lower than the results of the first setting.
The recognition performances largely deteriorated in this data-scarce setting even when data from the target subject was included in the training set.
Interestingly, F1-measures of U-Net and DCL-SA were lower than CNN and DCL, respectively, in the data-scarce setting.
A model with large trainable parameters or a self-attention module is likely to require more training data.
Therefore, methods should be developed to facilitate training of such state-of-the-art modules and architectures, such as self-attention, with limited training data \cite{morales2022acceleration}.

\subsubsection{Cross-user Activity Recognition with Different Amount of Training Data}
\label{subsubsec:benchmark3}

We investigate the relationship between the amount of training data used and recognition accuracy.
We used a fixed test set and the model was trained by varying the number of remaining workers.
Data from four workers, i.e., U0104, U0108/U0203, U0110/U0110, and U0207, was used as a test set.

\begin{figure}[t]
  \centering
  \vspace{-3mm}
  \includegraphics[width=1.0\linewidth]{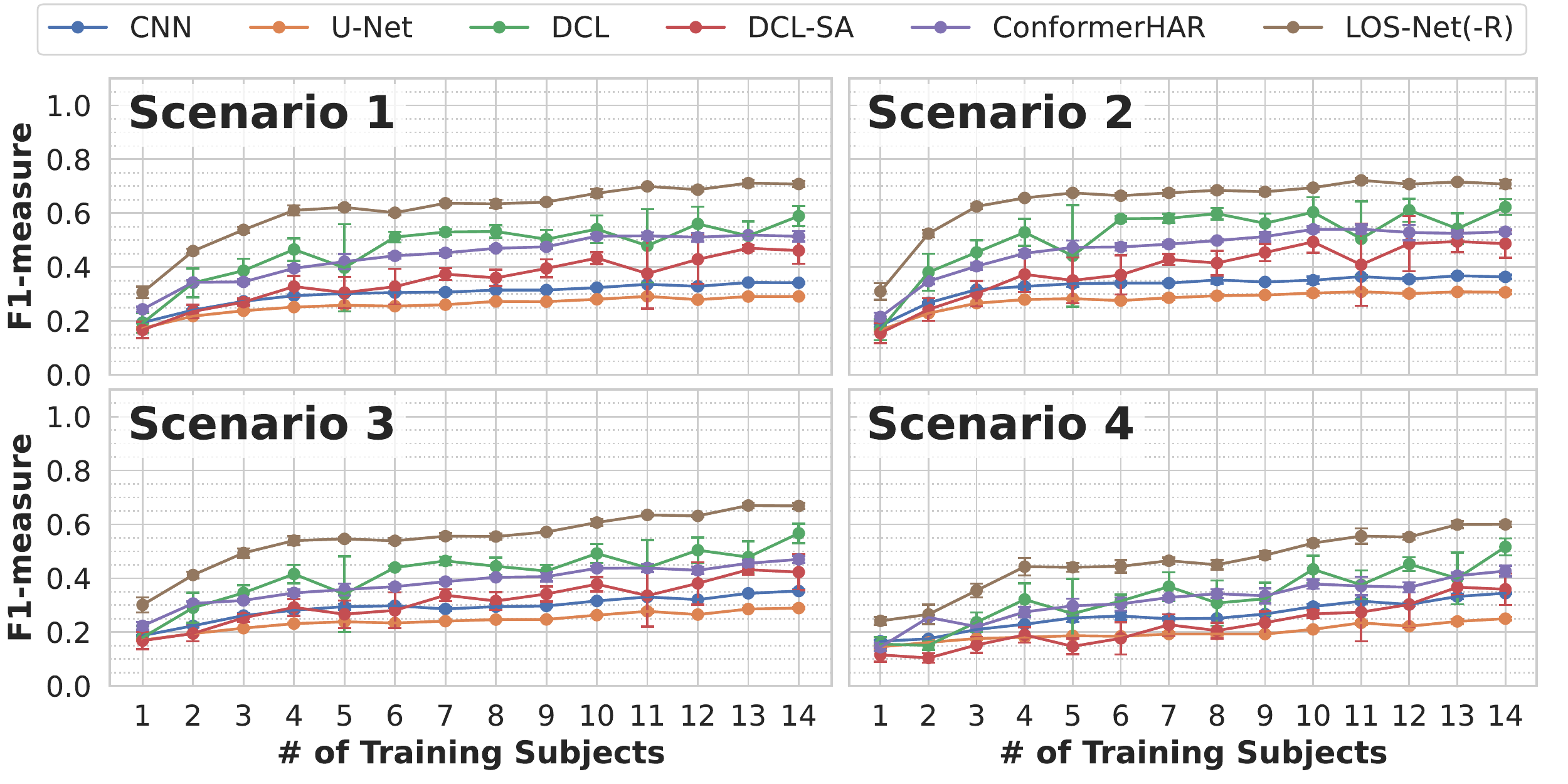}
  \vspace{-6mm}
  \caption{
    Results of work activity recognition with a limited amount of training data from source workers
  }
  \label{fig:f1_b3}
\end{figure}

Fig.~\ref{fig:f1_b3} shows the results.
In Scenario~1 and Scenario~2, recognition performance improved significantly until the number of training subjects reached a total of 4, after which performance improvements became moderate.
In contrast, in Scenarios~3 and Scenario~4, the score gradually improved with more training subjects, indicating the negative effect of variations in sensor data on recognition performance in these scenarios.
Therefore, the development of methods that can deal with variations in the data is crucial.
For example, for Scenario~4, in which the subjects were rushed, a model that is robust against differences in working speed must be developed, such as using data augmentation or a bottom-up approach that detects actions that are less sensitive to the differences in speed first and then estimates work operations.

\subsubsection{Recognition with Multi-/limited-modalities}
\label{subsubsec:benchmark4}

Multi-modal activity recognition is a hot research topic and this approach has the potential to achieve precise recognition compared to single-modal recognition.
However, because it is not practical to ask workers to wear multiple sensors on a daily basis, recognition technologies using limited modalities are an important research focus area.
Here, we evaluated the recognition performance of the various sensor combinations.
We used the setting as Section \ref{subsubsec:benchmark1}, i.e., leave-one-subject-out CV.
LOS-Net(-R) was used for wearable sensor modalities with early fusion techniques \cite{munzner2017cnn} and ST-GCN \cite{yan2018spatial} was used for keypoints.

Figure \ref{fig:f1_b4} shows the results.
Since most subjects are right-handed, the recognition accuracy of the combination including the right wrist sensor is higher than that of the other combinations.
However, the addition of gyros and quaternions to capture more detailed hand movements resulted in limited improvement in accuracy with the simple early fusion techniques.
ST-GCN exhibited relatively lower performance, probably due to the occlusions of boxes.
Based on these results, we believe there is considerable room for performance improvement by sensor fusion.

%% file: body/50_Discussion.tex
\section{Research Directions and Potential Tasks}
\label{subsec:research_task}

In this section, based on the analysis and benchmark results, we highlight the possible research directions that can be explored with OpenPack.

\noindent
\textbf{$\bullet$ Metadata-aided activity recognition}: The benchmark results showed that recognition performances for some models was low for activities affected by conditions such as the combination of items to pack. Because information about an order sheet that a worker is currently processing is commonly managed by an online order management system, the performance of activity recognition methods can be enhanced with information about the order such as the number and size of items. For example, we can switch activity recognition models depending on the characteristics of orders, and information about an order, such as the number of items, can be used as prior knowledge because the duration of related operations is proportional to the number of items.

\noindent
\textbf{$\bullet$ Speed-invariant activity recognition}: The working speed significantly affected recognition performance in the benchmark experiment. The working speed can vary between workers and depending on situations. Therefore, development of speed-invariant activity recognition is important, such as data augmentation techniques enabling recognition performance that is robust to variations in working speed and speed-agnostic feature extraction methods.

\noindent
\textbf{$\bullet$ Fusion with high-confidence modalities}: The class imbalance problem and sensor data variations are inevitable in complex work activity recognition, and result in performance deterioration for specific difficult activity classes.
Therefore, fusion with high-confidence modalities such as readings from IoT-enabled devices, such as a bar-code scanner, is crucial.

In addition to the operation and action recognition tasks using sensor data, OpenPack with a set of rich metadata and annotations enables various designs of research tasks, which includes the following: 
(1) transfer learning across sensor positions, across subjects, and across modalities \cite{kong2019mmact}, %
(2) skill assessment using sensor data and metadata related to work experience as ground truth \cite{xia2022comparative},
(3) counting the number of necessary actions or the number of packed items using sensor data \cite{nishino2021weakcounter},
(4) estimating workers’ levels of fatigue using sensor and physiological data,
and (5) detecting mistakes and accidents in the work process \cite{sener2022assembly101}.

\begin{figure}[t]
  \begin{center}
  \vspace{-3mm}
  \includegraphics[width=1.00\linewidth]{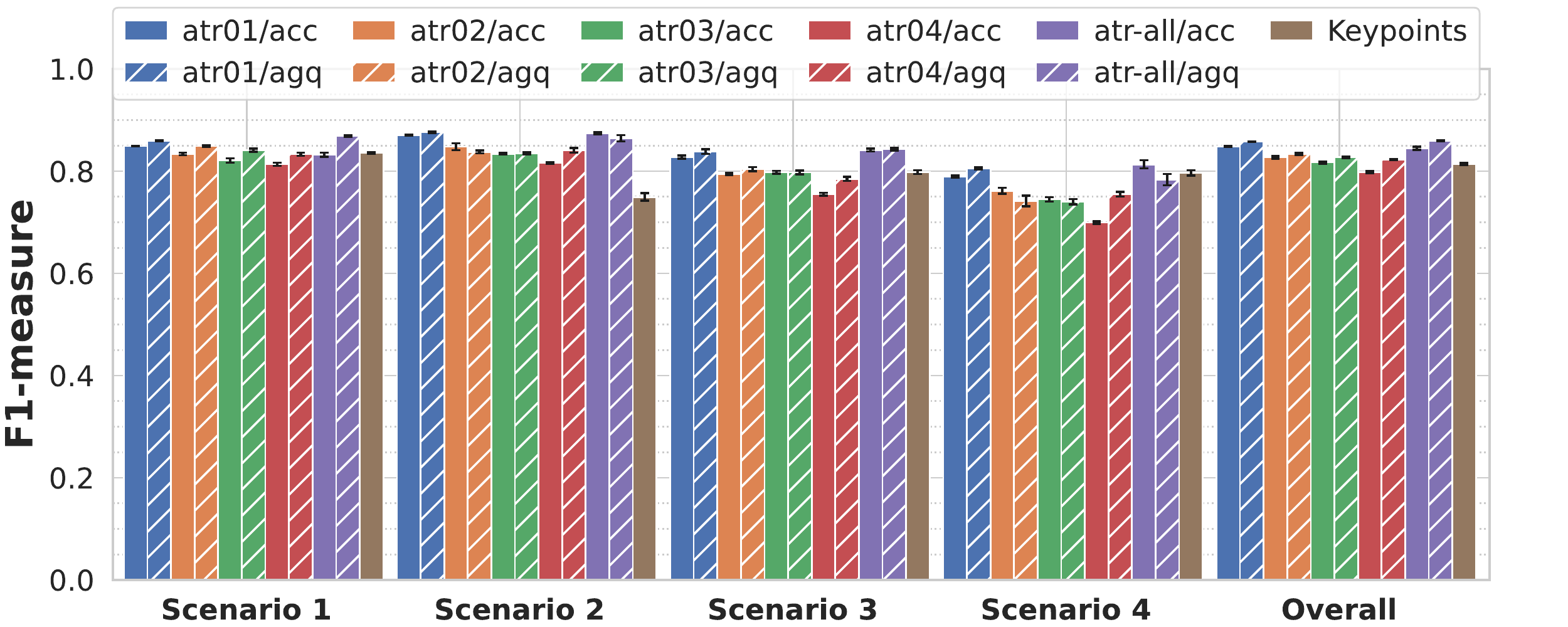}
  \vspace{-5mm}
  \caption{
    Results of work operation recognition with various sensor combinations.
    ``atr01'' to ``atr04'' are IMU sensors. 
    The input modality is acceleration only for "/acc" and a combination of acceleration, gyro, and quaternion for "/agq".
  }
  \label{fig:f1_b4}
\end{center}
\end{figure}

%% file: body/80_Conclusion.tex
\section{Conclusion}

This study presented a new large-scale dataset for packaging work recognition called OpenPack dataset.
Based on the analysis and benchmark results on OpenPack, we provided future research directions for complex work activity recognition.
We believe that human activity recognition methods developed based on OpenPack are applicable to many complex work activities in industrial domains,
and OpenPack can be used as a baseline dataset for complex work activity recognition.